\documentclass[twoside,11pt]{article}

\usepackage{times}
\usepackage{fullpage}

\usepackage{graphicx} 
\usepackage{subfigure}

\usepackage[round]{natbib}

\usepackage{algorithm}
\usepackage{algorithmic}
\usepackage{hyperref}

\usepackage{multirow}
\usepackage{enumitem}
\usepackage{xspace}
\usepackage{amssymb,amsfonts,latexsym,amsmath,amsthm}
\usepackage{textcomp}

\theoremstyle{plain}
\newtheorem{definition}{Definition}

\newtheorem{theorem}{Theorem}

\newcommand{\oursglobal}{SCML-Global\xspace}
\newcommand{\ourslocal}{SCML-Local\xspace}
\newcommand{\mtours}{mt-SCML\xspace}
\newcommand{\stours}{st-SCML\xspace}
\newcommand{\uours}{u-SCML\xspace}
\newcommand{\LMNNF}{Global-Frob\xspace}
\newcommand{\LMNN}{LMNN\xspace}
\newcommand{\PLML}{PLML\xspace}
\newcommand{\mLMNN}{MM-LMNN\xspace}
\newcommand{\mtLMNN}{mt-LMNN\xspace}
\newcommand{\stLMNN}{st-LMNN\xspace}
\newcommand{\uLMNN}{u-LMNN\xspace}
\newcommand{\boost}{BoostML\xspace}
\newcommand{\Euc}{Euc\xspace}
\newcommand{\stEuc}{st-Euc\xspace}
\newcommand{\uEuc}{u-Euc\xspace}

\newcommand{\GLML}{GLML\xspace}

\usepackage{amsmath,amsfonts}
\usepackage{amsthm,amssymb,amsopn}
\usepackage{bm} 

\newcommand{\vct}[1]{\boldsymbol{#1}} 
\newcommand{\mat}[1]{\boldsymbol{#1}} 

\newcommand{\T}{^{\textrm T}} 

\newcommand{\eat}[1]{}

\usepackage[title]{appendix}

\graphicspath{ {../} }

\begin{document}

\title{Sparse Compositional Metric Learning\thanks{This document is an extended version of a conference paper \citep{Shi2014} that provides additional details and results.}}
\author{
Yuan Shi\thanks{Equal contribution.}~~\thanks{Department of Computer Science, University of Southern California,
\texttt{\{yuanshi,bellet,feisha\}@usc.edu}.}~, Aur\'elien Bellet\footnotemark[2]~~\footnotemark[3]~, Fei Sha\footnotemark[3]
}

\date{}

\maketitle

\begin{abstract}
We propose a new approach for metric learning by framing it as
learning a sparse combination of locally discriminative metrics that
are inexpensive to generate from the training data. This flexible
framework allows us to naturally derive formulations for
global, multi-task and local metric learning. The resulting algorithms have several advantages over existing methods in the
literature: a much smaller number of parameters to be estimated and
a principled way to generalize learned metrics to new testing data
points. To analyze the approach theoretically, we derive a
generalization bound that justifies the sparse combination.
Empirically, we evaluate our algorithms on several datasets against
state-of-the-art metric learning methods. The results are consistent
with our theoretical findings and demonstrate the superiority of our
approach in terms of classification performance and
scalability.
\end{abstract}

\section{Introduction}

The need for measuring  distance or similarity between data instances is ubiquitous in machine learning and many application domains. However, each problem has its own underlying semantic space for defining distances that standard metrics (e.g., the Euclidean distance) often fail to capture. This has led to a growing interest in \textit{metric learning} for the past few years, as summarized in two recent surveys \citep{Bellet2013,Kulis2012}. Among these methods, learning a globally linear Mahalanobis distance 
is by far the most studied setting. Representative methods include \citep{Xing2002,Goldberger2004,Davis2007,Jain2008,Weinberger2009,Shen2012,Ying2012}. 
This is equivalent to learning a linear projection of the data to a feature space where constraints on the training set (such as ``$\vct{x}_i$ should be closer to $\vct{x}_j$ than to $\vct{x}_k$'') are better satisfied.

Although the performance of these learned metrics is typically superior to that of standard metrics in practice, a single linear metric is often unable to accurately capture the complexity of the task, for instance when the data are multimodal or the decision boundary is complex. To overcome this limitation, recent work has focused on learning \textit{multiple locally linear metrics} at several locations of the feature space \citep{Frome2007,Weinberger2009,Zhan2009,Hong2011,Wang2012b}, to the extreme of learning one metric per training instance~\citep{Noh2010}. This line of research is motivated by the fact that locally, simple linear metrics perform well \citep{Ramanan2011,Hauberg2012}. The main challenge is to integrate these metrics into a meaningful global one while keeping the number of learning parameters to a reasonable level in order to avoid heavy computational burden and severe overfitting. So far, existing methods are not able to compute valid (smooth) global metrics from the local metrics they learn and do not provide a principled way of generalizing to new regions of the space at test time. Furthermore, they scale poorly with the dimensionality $D$ of the data: typically, learning a Mahalanobis distance requires $O(D^2)$ parameters and the optimization involves projections onto the positive semidefinite cone that scale in $O(D^3)$. This is expensive even for a single metric when $D$ is moderately large.


\begin{figure}[t]
\centering
\includegraphics[width=0.95\textwidth]{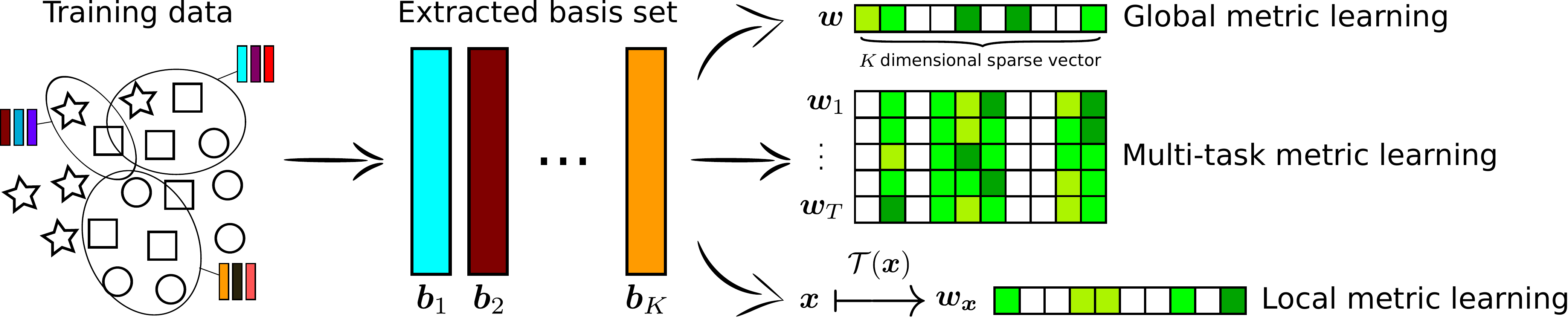}
\caption{Illustration of the general framework and its applications. We extract locally discriminative basis elements from the training data and cast metric learning as learning sparse combinations of these elements. We formulate global metric learning as learning a single sparse weight vector $\vct{w}$. For multi-task metric learning, we learn a vector $\vct{w}_t$ for each task where all tasks share the same basis subset. For local metric learning we learn a function $\mathcal{T}(\vct{x})$ that maps any instance $\vct{x}$ to its associated sparse weight vector $\vct{w}_{\vct{x}}$. Shades of grey encode weight magnitudes.}
\label{fig:idea}
\end{figure}

In this paper, we study metric learning from a new perspective to efficiently address these key challenges. We propose to learn metrics as
\textit{sparse compositions of locally discriminative metrics}. These ``basis metrics'' are low-rank and extracted efficiently from the training data at different local regions, for instance using Fisher discriminant analysis.
Learning higher-rank linear metrics is then formulated as learning the combining weights, using sparsity-inducing regularizers to select only the most useful basis elements.
This provides a unified framework for metric learning, as illustrated in Figure~\ref{fig:idea}, that we call  SCML (for Sparse Compositional Metric Learning). In SCML, the number of parameters to learn is much smaller than existing approaches and projections onto the positive semidefinite cone are not needed. This gives an efficient and flexible way to learn a single global metric when $D$ is large.

The proposed framework also applies to multi-task metric learning, where one wants to learn a global metric for several related tasks while exploiting commonalities between them~\citep{Caruana1997,Parameswaran2010}. This is done in a natural way by means of a group sparsity regularizer that makes the task-specific metrics share the same basis subset.
Our last and arguably most interesting contribution is a new formulation for local metric learning, where we learn a transformation $\mathcal{T}(\vct{x})$ that takes as input any instance $\vct{x}$ and outputs a sparse weight vector defining its metric. This can be seen as learning a smoothly varying metric tensor over the feature space \citep{Ramanan2011,Hauberg2012}. To the best of our knowledge, it is the first discriminative metric learning approach capable of computing, in a principled way, an instance-specific metric for \textit{any} point in the feature space. All formulations can be solved using scalable optimization procedures based on stochastic subgradient descent with proximal operators \citep{Duchi2009a,Xiao2010}.

We present both theoretical and experimental evidence supporting the proposed approach. We derive a generalization bound which provides a theoretical justification to seeking sparse combinations and suggests that the basis set $B$ can be large without incurring overfitting. Empirically, we evaluate our algorithms against
state-of-the-art global, local and multi-task metric learning methods on several datasets. The results strongly support the proposed framework. 

The rest of this paper is organized as follows. Section~\ref{sec:approach} describes our general framework and illustrates how it can be used to derive efficient formulations for global, local and multi-task metric learning. Section~\ref{sec:gen} provides a theoretical analysis supporting our approach. Section~\ref{sec:related} reviews related work. Section~\ref{sec:exp} presents an experimental evaluation of the proposed methods. We conclude in Section~\ref{sec:disc}.

\section{Proposed Approach}
\label{sec:approach}

In this section, we present the main idea of sparse compositional metric learning (SCML) and show how it can be used to unify several existing metric learning paradigms and lead to efficient new formulations.

\subsection{Main Idea}

We assume the data lie in $\mathbb{R}^D$ and focus on learning (squared) Mahalanobis distances $d_{\mat{M}}(\vct{x},\vct{x}') = (\vct{x}-\vct{x}')\T\mat{M}(\vct{x}-\vct{x}')$ parameterized by a positive semidefinite (PSD) $D\times D$ matrix $\mat{M}$. Note that $\mat{M}$ can be represented as a nonnegative weighted sum of $K$ rank-1 PSD matrices:\footnote{Such an expression exists for any PSD matrix $\mat{M}$ since the eigenvalue decomposition of $\mat{M}$ is of the form \eqref{eq:comb}.}
\begin{equation}
\label{eq:comb}
\mat{M} = \sum_{i=1}^Kw_i\vct{b}_i\vct{b}_i\T,\quad\text{with }\vct{w} \geq 0,
\end{equation}
where the $\vct{b}_i$'s are $D$-dimensional column vectors.

In this paper, we use the form \eqref{eq:comb} to cast metric learning as learning a \textit{sparse combination of basis elements} taken from a basis set $B = \{\vct{b}_i\}_{i=1}^K$. The key to our framework is the fact that such a $B$ is made readily available to the algorithm and consists of rank-one metrics that are {\it locally discriminative}. Such basis elements can be easily generated from the training data at several local regions --- in the experiments, we simply use Fisher discriminant analysis (see the corresponding section for details). They can then be combined to form a single global metric, multiple global metrics (in the multi-task setting) or a metric tensor (implicitly defining an infinite number of local metrics) that varies smoothly across the feature space, as we will show in later sections.

We use the notation $d_{\vct{w}}(\vct{x},\vct{x}')$ to highlight our parameterization of the Mahalanobis distance by $\vct{w}$. Learning $\mat{M}$ in this form makes it PSD by design (as a nonnegative sum of PSD matrices) and involves $K$ parameters (instead of $D^2$ in most metric learning methods), enabling it to more easily deal with high-dimensional problems. We also want the combination to be \textit{sparse}, i.e., some $w_i$'s are zero and thus $\mat{M}$ only depends on a small subset of $B$. This provides some form of regularization (as shown later in Theorem~\ref{thm:bound}) as well as a way to tie metrics together when learning multiple metrics.
In the rest of this section, we apply the proposed framework to several metric learning paradigms (see Figure~\ref{fig:idea}). We start with the simple case of global metric learning (Section~\ref{sec:global}) before considering more challenging settings: multi-task (Section~\ref{sec:mt}) and local metric learning (Section~\ref{sec:local}). Finally, Section~\ref{sec:optimization} discusses how these formulations can be solved in a scalable way using stochastic subgradient descent with proximal operators.

\subsubsection{Global Metric Learning}
\label{sec:global}

In global metric learning, one seeks to learn a single metric $d_{\vct{w}}(\vct{x},\vct{x}')$ from a set of distance constraints on the training data. Here, we use a set of triplet constraints $C$ where each $(\vct{x}_i,\vct{x}_j,\vct{x}_k)\in C$ indicates that the distance between $\vct{x}_i$ and $\vct{x}_j$ should be smaller than the distance between $\vct{x}_i$ and $\vct{x}_k$. $C$ may be constructed from label information, as in LMNN \citep{Weinberger2009}, or in an unsupervised manner based for instance on implicit users' feedback (such as clicks on search engine results).
Our formulation for global metric learning, \oursglobal, is simply to combine the local basis elements into a higher-rank global metric that satisfies well the constraints in $C$:
\begin{equation}
\label{eq:global}
\min_{\vct{w}} \frac{1}{|C|}\sum_{(\vct{x}_i,\vct{x}_j,\vct{x}_k)\in C} L_{\vct{w}}(\vct{x}_i,\vct{x}_j,\vct{x}_k)+ \beta\|\vct{w}\|_1,
\end{equation}
where $L_{\vct{w}}(\vct{x}_i,\vct{x}_j,\vct{x}_k) = [ 1 + d_{\vct{w}}(\vct{x}_i,\vct{x}_j) - d_{\vct{w}}(\vct{x}_i,\vct{x}_k)]_+$ with $[\cdot]_+ = \max(0,\cdot)$, and $\beta \geq 0$ is a regularization parameter. The first term in \eqref{eq:global} is the classic margin-based hinge loss function. 
The second term $\|\vct{w}\|_1 = \sum_{i=1}^K w_i$ is the $\ell_1$ norm regularization which encourages sparse solutions, allowing the selection of relevant basis elements. \oursglobal is convex by the linearity of both terms and is bounded below, thus it has a global minimum.

\subsubsection{Multi-Task Metric Learning}
\label{sec:mt}

Multi-task learning \citep{Caruana1997} is a paradigm for learning
several tasks simultaneously, exploiting their
commonalities. When tasks are related, this can perform better than separately learning each task. Recently, multi-task
learning methods have successfully built on the assumption that the
tasks should share a common low-dimensional representation
\citep{Argyriou2008,Yang2009,Gong2012}. In general, it is unclear how to achieve this in metric learning. In contrast, learning metrics as sparse combinations allows a direct
translation of this idea to multi-task metric learning.

Formally, we are given
$T$ different but somehow related tasks with associated constraint
sets $C_1,\dots,C_T$ and we aim at learning a metric
$d_{\vct{w}_t}(\vct{x},\vct{x}')$ for each task $t$ while sharing
information across tasks. In the following, the basis set $B$ is the union of the basis sets $B_1,\dots,B_T$ extracted from each task $t$. Our formulation for multi-task metric
learning, \mtours, is as follows:
\begin{equation*}
\min_{\mat{W}}
\displaystyle\sum_{t=1}^T\frac{1}{|C_t|}\sum_{(\vct{x}_i,\vct{x}_j,\vct{x}_k)\in
C_t} L_{\vct{w}_t}(\vct{x}_i,\vct{x}_j,\vct{x}_k) +
\beta\|\mat{W}\|_{2,1},
\end{equation*}
where $\mat{W}$ is a $T\times K$ nonnegative matrix whose $t$-th row
is the weight vector $\vct{w}_t$ defining the metric for task $t$, $L_{\vct{w}_t}(\vct{x}_i,\vct{x}_j,\vct{x}_k) = \left[ 1 + d_{\vct{w}_t}(\vct{x}_i,\vct{x}_j) -
d_{\vct{w}_t}(\vct{x}_i,\vct{x}_k)\right]_+$
and $\|\mat{W}\|_{2,1}$ is the $\ell_2/\ell_1$ mixed norm used in
the group lasso problem \citep{Yuan2006}. It corresponds to the
$\ell_1$ norm applied to the $\ell_2$ norm of the columns of
$\mat{W}$ and is known to induce group sparsity at the column level. In other words, this regularization makes most basis elements either have zero weight or nonzero weight {\it for all tasks}.

Overall, while each metric remains task-specific ($d_{\vct{w}_t}$ is only required to satisfy well the constraints in $C_t$), it is composed of {\it shared features} (i.e., it potentially benefits from basis elements generated from other tasks) that are regularized to be relevant \textit{across tasks} (as favored by the group sparsity). As a result, all learned metrics can be expressed as combinations of the same basis subset of $B$, though with different weights for each task. Since the $\ell_2/\ell_1$ norm is
convex, \mtours is again convex.


\subsubsection{Local Metric Learning}
\label{sec:local}

Local metric learning addresses the limitations of global methods in capturing complex data patterns \citep{Frome2007,Weinberger2009,Zhan2009,Noh2010,Hong2011,Wang2012b}. For heterogeneous data,  allowing the metric to vary across the feature space can capture the semantic distance much better. On the other hand, local metric learning is costly and often suffers from severe overfitting since the number of parameters to learn can be very large. In the following, we show how our framework can be used to derive an efficient local metric learning method.

We aim at learning a \textit{metric tensor} $\mathcal{T}(\vct{x})$, which is a smooth function that (informally) maps any instance $\vct{x}$ to its metric matrix  \citep{Ramanan2011,Hauberg2012}. The distance between two points should then be defined as the geodesic distance on a Riemannian manifold. However, this requires solving an intractable problem, so we use the widely-adopted simplification that distances from point $\vct{x}$ are computed based on its own metric alone \citep{Zhan2009,Noh2010,Wang2012b}:
\begin{eqnarray*}
d_{\mathcal{T}}(\vct{x},\vct{x}') &=& (\vct{x}-\vct{x}')\T \mathcal{T}(\vct{x})(\vct{x}-\vct{x}')\\
&=& (\vct{x}-\vct{x}')\T \sum_{i=1}^Kw_{\vct{x},i}\vct{b}_i\vct{b}_i\T(\vct{x}-\vct{x}'),
\end{eqnarray*}
where $\vct{w}_{\vct{x}}$ is the weight vector for instance
$\vct{x}$.

We could learn a weight vector for each
training point. This would result in a formulation similar to \mtours, where each training
instance is considered as a task. However, in the context of local
metric learning, this is not an appealing solution. Indeed, for a
training sample of size $S$ we would need to learn $SK$ parameters, which
is computationally difficult and leads to heavy overfitting for
large-scale problems. Furthermore, this gives no principled way of
computing the weight vector of a test instance.

We instead propose a
more effective solution by constraining the weight vector for an
instance $\vct{x}$ to parametrically depend on some embedding of
$\vct{x}$:
\begin{equation}
\label{eq:tensor} \mathcal{T}_{\mat{A},\vct{c}}(\vct{x}) =
\sum_{i=1}^K(\vct{a_i}\T\vct{z}_{\vct{x}} + c_i)^2\vct{b}_i\vct{b}_i\T,
\end{equation}
where $\vct{z}_{\vct{x}} \in D'$ is an embedding of $\vct{x}$,\footnote{In our experiments, we use kernel PCA \citep{Scholkopf1998} as it provides a simple way to limit the dimension and thus the number of parameters to learn. We use RBF kernel with bandwidth set to the median Euclidean distance in the data.} $\mat{A} =
[\vct{a_1} \dots \vct{a_K}]\T$ is a $D'\times K$ real-valued matrix
and $\vct{c} \in \mathbb{R}^K$. The square makes the weights
nonnegative $\forall \vct{x}\in \mathbb{R}^D$, ensuring that they
define a valid (pseudo) metric. Intuitively, \eqref{eq:tensor} combines the locally discriminative metrics with weights that depend on the position of the instance in the feature space.

There are several advantages to this
formulation. First, 
by learning
$\mat{A}$ and $\vct{c}$ we implicitly learn a different metric not
only for the training data but for any point in the feature space. Second, if the embedding is smooth,
$\mathcal{T}_{\mat{A},\vct{c}}(\vct{x})$ is a smooth function of $\vct{x}$,
therefore similar instances are assigned similar weights. This can
be seen as some kind of manifold regularization. Third, the number
of parameters to learn is now $K(D'+1)$, thus independent of
both the size of the training sample and the dimensionality of $\vct{x}$.
Our formulation for local metric learning, \ourslocal, is as
follows:
\begin{equation*}
\min_{\tilde{\mat{A}}} \frac{1}{|C|}\sum_{(\vct{x}_i,\vct{x}_j,\vct{x}_k)\in C} L_{\mathcal{T}_{\mat{A},\vct{c}}}(\vct{x}_i,\vct{x}_j,\vct{x}_k) + \beta\|\tilde{\mat{A}}\|_{2,1},
\end{equation*}
where $\tilde{\mat{A}}$ is a $(D' + 1)\times K$ matrix denoting the concatenation of $\mat{A}$ and $\vct{c}$, and $L_{\mathcal{T}_{\mat{A},\vct{c}}}(\vct{x}_i,\vct{x}_j,\vct{x}_k) = \left[ 1 + d_{\mathcal{T}_{\mat{A},\vct{c}}}(\vct{x}_i,\vct{x}_j) - d_{\mathcal{T}_{\mat{A},\vct{c}}}(\vct{x}_i,\vct{x}_k)\right]_+$. The $\ell_2/\ell_1$ norm on $\tilde{\mat{A}}$ introduces sparsity at the column level, regularizing the local metrics to use the same basis subset. Interestingly, if $\mat{A}$ is the zero matrix, we recover
\oursglobal. \ourslocal is nonconvex 
and is thus subject to local minima.


\subsection{Optimization}
\label{sec:optimization}

Our formulations use (nonsmooth) sparsity-inducing regularizers and typically involve a large number of triplet constraints. We can solve them efficiently using stochastic composite optimization \citep{Duchi2009a,Xiao2010}, which alternates between a stochastic subgradient step on the hinge loss term and a proximal operator (for $\ell_1$ or $\ell_{2,1}$ norm) that explicitly induces sparsity. We solve \oursglobal and \mtours using Regularized Dual Averaging \citep{Xiao2010}, which offers fast convergence and levels of sparsity in the solution comparable to batch algorithms. For \ourslocal, due to  local minima, we ensure improvement over the optimal solution $\vct{w}^*$ of \oursglobal by using a forward-backward algorithm \citep{Duchi2009a} which is initialized with $\mat{A} = \mat{0}$ and $c_i = \sqrt{w^*_i}$.

Recall that unlike most existing metric learning algorithms, we do not need to perform projections onto the PSD cone, which scale in $O(D^3)$ for a $D\times D$ matrix. Our algorithms thereby have a significant computational advantage for high-dimensional problems.

%

\section{Theoretical Analysis}
\label{sec:gen}

In this section, we provide a theoretical analysis of our approach in the form of a generalization bound based on algorithmic robustness analysis \citep{Xu2012a} and its adaptation to metric learning \citep{Bellet2012b}. For simplicity, we focus on \oursglobal, our global metric learning formulation in \eqref{eq:global}.

Consider the supervised learning setting, where we are given a labeled training sample $S=\{\vct{z}_i=(\vct{x}_i,y_i)\}_{i=1}^n$ drawn i.i.d. from some unknown distribution $P$ over $\mathcal{Z} = \mathcal{X}\times \mathcal{Y}$. We call a triplet $(\vct{z},\vct{z}',\vct{z}'')$ \textit{admissible} if $y = y' \neq y''$.
Let $C$ be the set of admissible triplets built from $S$ and $L(\vct{w},\vct{z},\vct{z}',\vct{z}'') = \left[ 1 + d_{\vct{w}}(\vct{x},\vct{x}') - d_{\vct{w}}(\vct{x},\vct{x}'')\right]_+$ denote the loss function used in \eqref{eq:global}, with the convention that $L$ returns 0 for non-admissible triplets.

Let us define the {\it empirical loss} of $\vct{w}$ on $S$ as
$$\mathcal{R}_{emp}^S(\vct{w}) = \frac{1}{|C|}\sum_{(\vct{z},\vct{z}',\vct{z}'')\in C} L(\vct{w},\vct{z},\vct{z}',\vct{z}''),$$
and its {\it expected loss} over distribution $P$ as
$$\mathcal{R}(\vct{w}) = \mathbb{E}_{\vct{z},\vct{z}',\vct{z}''\sim P} L(\vct{w},\vct{z},\vct{z}',\vct{z}'').$$

The following theorem bounds the deviation between the empirical loss of the learned metric and its expected loss.

\begin{theorem}
\label{thm:bound}
Let $\vct{w}^*$ be the optimal solution to \oursglobal with $K$ basis elements, $\beta>0$ and $C$ constructed from $S=\{(\vct{x}_i,y_i)\}_{i=1}^n$ as above.  Let $K^*\leq K$ be the number of nonzero entries in $\vct{w}^*$. Let us assume the norm of any instance bounded by some constant $R$ and $L$ uniformly upper-bounded by some constant $U$. Then
for any $\delta>0$, with probability at least $1-\delta$ we have:
$$
\left|\mathcal{R}(\vct{w}^*)-\mathcal{R}_{emp}^S(\vct{w}^*)\right|\leq
\frac{16\gamma RK^*}{\beta}+3U\sqrt{\frac{N \ln 2 + \ln
\frac{1}{\delta}}{0.5n}},
$$
where $N$ is the size of an $\gamma$-cover of $\mathcal{Z}$.
\end{theorem}

This bound has a standard $O(1/\sqrt{n})$ asymptotic convergence rate.\footnote{In robustness bounds, the cover radius $\gamma$ can be made arbitrarily close to zero at the expense of increasing $N$. Since $N$ appears in the second term, the right hand side of the bound indeed goes to zero when $n\rightarrow \infty$. This is in accordance with other similar learning bounds, for example, the original robustness-based bounds in \citep{Xu2012a}.} Its main originality is that it provides a theoretical justification to enforcing sparsity in our formulation. Indeed, notice that $K^*$ (and not $K$) appears in the bound as a penalization term, which suggests that one may use a large basis set $K$ without overfitting as long as $K^*$ remains small. This will be confirmed by our experiments (Section~\ref{sec:explocal}). A similar bound can be derived for \mtours, but not for \ourslocal because of its nonconvexity. The details and proofs can be found in Appendix~\ref{app:analysis}.

\section{Related Work}
\label{sec:related}

In this section, we review relevant work in global, multi-task and local metric learning. The interested reader should refer to the recent surveys of \citet{Kulis2012} and \citet{Bellet2013} for more details.

\paragraph{Global methods}
Most global metric learning methods learn the matrix $\mat{M}$ directly: see \citep{Xing2002,Goldberger2004,Davis2007,Jain2008,Weinberger2009} for representative papers. This is computationally expensive and subject to overfitting for moderate to high-dimensional problems.
An exception is BoostML \citep{Shen2012} which uses rank-one matrices as weak learners to learn a global Mahalanobis distance via a boosting procedure. 
However, it is not clear how BoostML can be generalized to multi-task or local metric learning.

\paragraph{Multi-task methods}
Multi-task metric learning was proposed in \citep{Parameswaran2010} as an extension to the popular LMNN \citep{Weinberger2009}. The authors define the metric for task $t$ as $d_t(\vct{x},\vct{x}') = (\vct{x}-\vct{x}')\T(\mat{M}_0 + \mat{M}_t)(\vct{x}-\vct{x}')$, where $\mat{M}_t$ is task-specific and $\mat{M}_0$ is shared by all tasks. Note that it is straightforward to incorporate their approach in our framework by defining a shared weight vector $\vct{w}_0$ and task-specific weights $\vct{w}_t$.
However, this assumption of a metric that is common to all tasks can be too restrictive in cases where task relatedness is complex, as illustrated by our experiments.

\paragraph{Local methods}
MM-LMNN \citep{Weinberger2009} is an extension of LMNN which learns only a small number of metrics (typically one per class) in an effort to alleviate overfitting. However, no additional regularization is used and a full-rank metric is learned for each class, which becomes intractable when the number of classes is large.
msNCA \citep{Hong2011} learns a function that splits the space into a small number of regions and then learns a metric per region using NCA \citep{Goldberger2004}. Again, the metrics are full-rank so msNCA does not scale well with the number of metrics.
Like \ourslocal, PLML \citep{Wang2012b} is based on a combination of metrics but there are major differences with our work: (i) weights only depend on a manifold assumption: they are not sparse and use no discriminative information, (ii) the basis metrics are full-rank, thus expensive to learn, and (iii) a weight vector is learned explicitly for each training instance, which can result in a large number of parameters and prevents generalization to new instances (in practice, for a test point, they use the weight vector of its nearest neighbor in the training set).
As observed by \citet{Ramanan2011}, the above methods make the implicit assumption that the metric tensor is locally constant (at the class, region or neighborhood level), while \ourslocal learns a smooth function that maps any instance to its specific metric.
ISD \citep{Zhan2009} is an attempt to learn the metrics for unlabeled points by propagation, but is limited to the transductive setting. Unlike the above discriminative approaches, GLML \citep{Noh2010} learns a metric for each point independently in a generative way by minimizing the 1-NN expected error under some assumption for the class distributions.

\section{Experiments}
\label{sec:exp}

In this section, we compare our methods to state-of-the-art
algorithms on global, multi-task and local metric learning.\footnote{For all compared methods we use {\sc Matlab}
code from the authors' website. The {\sc Matlab} code for our methods is available at \url{http://www-bcf.usc.edu/~bellet/}.} We use a 3-nearest neighbor classifier in all experiments. To generate a set of locally
discriminative rank-one metrics, we first divide data into
regions via clustering. For each region center, we select $J$
nearest neighbors from each class (for $J=\{10,20,50\}$ to
account for different scales), and apply Fisher discriminant
analysis followed by eigenvalue decomposition to obtain the basis
elements.\footnote{We also experimented with a basis set based on local GLML metrics. Preliminary results were comparable to those obtained with the procedure above.
}
Section~\ref{sec:expglobal} presents results for global metric learning, Section~\ref{sec:expmulti} for multi-task and Section~\ref{sec:explocal} for local metric learning.

\subsection{Global Metric Learning}
\label{sec:expglobal}

\begin{table}[t]
\centering
\small{ \tabcolsep=0.13cm
\begin{tabular}{|l|c|c|c|c|c|c|}
\hline & Vehicle & Vowel & Segment & Letters & USPS & BBC\\
\hline \# samples & 846 & 990 & 2,310  & 20,000 & 9,298& 2,225\\
\hline \# classes & 4 & 11 & 7  & 26 & 10& 5\\
\hline \# features & 18 & 10 & 19  & 16 & 256& 9,636\\
\hline
\end{tabular}}
\caption{Datasets for global and local metric learning.}
\label{tab:data}
\end{table}

We use 6 datasets from UCI\footnote{\url{http://archive.ics.uci.edu/ml/}} and BBC\footnote{\url{http://mlg.ucd.ie/datasets/bbc.html}} (see Table~\ref{tab:data}). The dimensionality of USPS and BBC is reduced to 100 and 200 using PCA to speed up computation. We normalize the data as in \citep{Wang2012b} and split into train/validation/test (60\%/20\%/20\%), except for Letters and USPS where we use 3,000/1,000/1,000. Results are averaged over 20 random splits.

\subsubsection{Proof of Concept}
\label{sec:globalproof}

\paragraph{Setup} Global metric learning is a convenient setting to study the effect of combining basis elements. To this end, we consider a formulation with the same loss function as \oursglobal but that directly learns the metric matrix, using Frobenius norm regularization to reduce overfitting. We refer to it as \LMNNF.
Both algorithms use the same training triplets, generated by
identifying 3 target neighbors (nearest neighbors with same label) and
10 imposters (nearest neighbors with different label) for each
instance. We tune the
regularization parameter on the validation data. For \oursglobal, we
use a basis set of 400 elements for Vehicle, Vowel, Segment and BBC,
and 1,000 elements for Letters and USPS.

\begin{table}[t]
\centering
\small{ \tabcolsep=0.13cm
\begin{tabular}{|c|c|c|c|}
\hline Dataset & \Euc & \LMNNF & \oursglobal \\
\hline\hline Vehicle &29.7$\pm$0.6& \textbf{21.5$\pm$0.8} & \textbf{21.3$\pm$0.6}\\
\hline Vowel &\textbf{11.1$\pm$0.4}& \textbf{10.3$\pm$0.4} & \textbf{10.9$\pm$0.5}\\
\hline Segment &5.2$\pm$0.2& \textbf{4.1$\pm$0.2} & \textbf{4.1$\pm$0.2}\\
\hline Letters &14.0$\pm$0.2& \textbf{9.0$\pm$0.2} & \textbf{9.0$\pm$0.2}\\
\hline USPS &10.3$\pm$0.2& 5.1$\pm$0.2 & \textbf{4.1$\pm$0.1}\\
\hline BBC &8.8$\pm$0.3& 5.5$\pm$0.3 & \textbf{3.9$\pm$0.2} \\
\hline
\end{tabular}}
\caption{Global metric learning results (best in bold).} \label{tab:global}
\end{table}

\paragraph{Results}  Table~\ref{tab:global} shows misclassification rates with standard errors, where \Euc is the Euclidean distance. 
The results show that \oursglobal performs similarly as \LMNNF on low-dimensional datasets but has a clear advantage when dimensionality is high (USPS and BBC). This demonstrates that
learning a sparse combination of basis elements is an effective way to reduce overfitting and improve generalization. \oursglobal is also faster to train than \LMNNF on these datasets  (about 2x faster on USPS and 3x on BBC) because it does not require any
PSD projection. 

\subsubsection{Comparison to Other Global Algorithms}

\paragraph{Setup} We now compare \oursglobal to two state-of-the-art global metric learning algorithms: Large Margin Nearest Neighbor \citep[\LMNN,][]{Weinberger2009} and \boost \citep{Shen2012}. The datasets, preprocessing and setting for \oursglobal are the same as in Section~\ref{sec:globalproof}. \LMNN uses 3 target neighbors and all imposters, while these are set to 3 and 10 respectively for \boost (as in \oursglobal). 

\begin{table}[t]
\centering
\small{ \tabcolsep=0.13cm
\begin{tabular}{|c|c|c|c|c|}
\hline Dataset & \Euc & \LMNN & \boost & \oursglobal\\
\hline\hline Vehicle &29.7$\pm$0.6& 23.5$\pm$0.7 & \textbf{19.9$\pm$0.6} & 21.3$\pm$0.6\\
\hline Vowel &\textbf{11.1$\pm$0.4}& \textbf{10.8$\pm$0.4} & 11.4$\pm$0.4 & \textbf{10.9$\pm$0.5}\\
\hline Segment &5.2$\pm$0.2& 4.6$\pm$0.2 & \textbf{3.8$\pm$0.2} & \textbf{4.1$\pm$0.2}\\
\hline Letters &14.0$\pm$0.2& 11.6$\pm$0.3 & 10.8$\pm$0.2 & \textbf{9.0$\pm$0.2} \\
\hline USPS &10.3$\pm$0.2& \textbf{4.1$\pm$0.1} & 7.1$\pm$0.2 & \textbf{4.1$\pm$0.1} \\
\hline BBC &8.8$\pm$0.3& \textbf{4.0$\pm$0.2} & 9.3$\pm$0.3 & \textbf{3.9$\pm$0.2} \\
\hline\hline
Avg. rank & 3.3 & 2.0 & 2.3 & \textbf{1.2}\\
\hline
\end{tabular}}
\caption{Comparison of \oursglobal against LMNN and BoostML (best in bold).}
\label{tab:globalsup}
\end{table}

\paragraph{Results} Table~\ref{tab:globalsup} shows the average misclassification rates, along with standard error and the average rank of each method across all datasets. \oursglobal clearly outperforms \LMNN and \boost, ranking first on 5 out of 6 datasets and achieving the overall highest rank. Furthermore, its training time is smaller than competing methods, especially for high-dimensional data. For instance, on the BBC dataset, \oursglobal trained in about 90 seconds, which is about 20x faster than \LMNN and 35x faster than \boost. Note also that \oursglobal is consistently more accurate than linear SVM, as shown in Appendix~\ref{app:svm}.

\begin{table}[t]
\centering
\small{ \tabcolsep=0.13cm
\begin{tabular}{|c|c|c|}
\hline Dataset & \boost & \oursglobal \\
\hline\hline Vehicle & 334 & 164 \\
\hline Vowel & 19 & 47 \\
\hline Segment & 442 & 49 \\
\hline Letters & 20 & 133 \\
\hline USPS & 2,375 & 300 \\
\hline BBC & 3,000 & 59 \\
\hline
\end{tabular}}
\caption{Average number of basis elements in the solution.}
\label{tab:nbasis}
\end{table}

\paragraph{Number of selected basis elements} Like \oursglobal, recall that \boost is based on combining rank-one elements (see Section~\ref{sec:related}). The main difference with \oursglobal is that our method is given a set of locally discriminative metrics and picks the relevant ones by learning sparse weights, while BoostML generates a new basis element at each iteration and adds it to the current combination. Table~\ref{tab:nbasis} reports the number of basis elements used in \oursglobal and \boost solutions. Overall, \oursglobal uses fewer elements than \boost (on two datasets, it uses more but this yields to significantly better performance). The results on USPS and BBC also suggest that the number of basis elements selected by \oursglobal seems to scale well
with dimensionality. These nice properties come from its knowledge
of the entire basis set and the sparsity-inducing regularizer. On
the contrary, the number of elements (and therefore iterations)
needed by \boost to converge seems to scale poorly with
dimensionality.

\subsection{Multi-task Metric Learning}
\label{sec:expmulti}

\begin{table}[t]
\centering
\small{
\begin{tabular}{|c||c|c|c||c|c|c||c|c|}
\hline Task & \stEuc & \stLMNN & \stours & \uEuc & \uLMNN & \uours
& \mtLMNN & \mtours \\
\hline\hline Books & 33.5$\pm$0.5 & 29.7$\pm$0.4 & 27.0$\pm$0.5 & 33.7$\pm$0.5 & 29.6$\pm$0.4 & 28.0$\pm$0.4 & 29.1$\pm$0.4 & 25.8$\pm$0.4 \\
\hline DVD & 33.9$\pm$0.5 & 29.4$\pm$0.5 & 26.8$\pm$0.4 & 33.9$\pm$0.5 & 29.4$\pm$0.5 & 27.9$\pm$0.5 & 29.5$\pm$0.5 & 26.5$\pm$0.5 \\
\hline Electronics & 26.2$\pm$0.4 & 23.3$\pm$0.4 & 21.1$\pm$0.5 & 29.1$\pm$0.5 & 25.1$\pm$0.4 & 22.9$\pm$0.4 & 22.5$\pm$0.4 & 20.2$\pm$0.5 \\
\hline Kitchen & 26.2$\pm$0.6 & 21.2$\pm$0.5 & 19.0$\pm$0.4 & 27.7$\pm$0.5 & 23.5$\pm$0.3 & 21.9$\pm$0.5 & 22.1$\pm$0.5 & 19.0$\pm$0.4 \\
\hline
\hline Avg. accuracy & 30.0$\pm$0.2 & 25.9$\pm$0.2 & 23.5$\pm$0.2 & 31.1$\pm$0.3 & 26.9$\pm$0.2 & 25.2$\pm$0.2 & 25.8$\pm$0.2 & \textbf{22.9$\pm$0.2} \\
\hline Avg. runtime & N/A & 57 min & 3 min & N/A & 44 min & 2 min & 41 min & 5 min \\
\hline
\end{tabular}}
\caption{Multi-task metric learning results.}
\label{tab:mt}
\end{table}

\paragraph{Dataset} 
Sentiment Analysis \citep{Blitzer2007} is a popular dataset for multi-task learning that consists of Amazon reviews on four product types: kitchen appliances, DVDs,
books and electronics. Each product type is treated as a task and
has 1,000 positive and 1,000 negative reviews. 
To reduce computational cost,
we represent each review by a 200-dimensional feature vector by
selecting top 200 words of the largest mutual information with the
labels. We randomly split the dataset into training (800
samples), validation (400 samples) and testing (400 samples) sets.

\paragraph{Setup}  We compare the following metrics: \stEuc (Euclidean distance), \stLMNN and \stours (single-task
LMNN and single-task \oursglobal, trained independently on each
task), \uEuc (Euclidean trained on the union of the training data
from all tasks), \uLMNN (\LMNN on union), \uours (\oursglobal on
union), multi-task LMNN \citep{Parameswaran2010} and
finally our own multi-task method \mtours. We tune the
regularization parameters in \mtLMNN, \stours, \uours and \mtours on
validation sets. As in the previous experiment, the number of target
neighbors and imposters for our methods are set to 3 and 10
respectively.
We use a basis set of 400 elements for each task for \stours, the
union of these (1,600) for \mtours, and 400 for \uours.

\paragraph{Results}  Table~\ref{tab:mt} shows the results averaged over 20 random splits.
First, notice that \uLMNN and \uours obtain significantly higher
error rates than \stLMNN and \stours respectively, which suggests
that the dataset may violate \mtLMNN's assumption that all tasks
share a similar metric. Indeed, \mtLMNN does not outperform \stLMNN
significantly. On the other hand, \mtours performs better than its
single-task counterpart and than all other compared methods by a
significant margin, demonstrating its ability to leverage some
commonalities between tasks that \mtLMNN is unable to capture. It is
worth noting that the solution found by \mtours is based on only 273
basis elements on average (out of a total of 1,600), while \stours
makes use of significantly more elements (347 elements \textit{per
task} on average). Basis elements selected by \mtours are evenly
distributed across all tasks, which indicates that it is able to
exploit meaningful information across tasks to get both more
accurate and more compact metrics. Finally, note that our algorithms are about an order of magnitude faster.


\subsection{Local Metric Learning}
\label{sec:explocal}

\begin{table}[t]
\centering
\small{ \tabcolsep=0.13cm
\begin{tabular}{|c|c|c|c|c|}
\hline Dataset & \mLMNN & \GLML & \PLML & \ourslocal \\
\hline\hline Vehicle & 23.1$\pm$0.6 & 23.4$\pm$0.6 & 22.8$\pm$0.7 & \textbf{18.0$\pm$0.6}\\
\hline Vowel & 6.8$\pm$0.3 & \textbf{4.1$\pm$0.4} & 8.3$\pm$0.4 & 6.1$\pm$0.4\\
\hline Segment & \textbf{3.6$\pm$0.2} & \textbf{3.9$\pm$0.2} & \textbf{3.9$\pm$0.2}& \textbf{3.6$\pm$0.2}\\
\hline Letters & 9.4$\pm$0.3 & 10.3$\pm$0.3 & \textbf{8.3$\pm$0.2} & \textbf{8.3$\pm$0.2}\\
\hline USPS & \textbf{4.2$\pm$0.7} & 7.8$\pm$0.2 & 4.1$\pm$0.1 & \textbf{3.6$\pm$0.1}\\
\hline BBC & 4.9$\pm$0.4 & 5.7$\pm$0.3 & \textbf{4.3$\pm$0.2} & \textbf{4.1$\pm$0.2}\\
\hline\hline
Avg. rank & 2.0 & 2.7 & 2.0 & \textbf{1.2} \\
\hline
\end{tabular}}
\caption{Local metric learning results (best in bold).} \label{tab:local}
\end{table}


\paragraph{Setup}  We use the same datasets and preprocessing as for global
metric learning. We compare \ourslocal to \mLMNN \citep{Weinberger2009}, \GLML
\citep{Noh2010} and \PLML
\citep{Wang2012b}.
The parameters of all methods are tuned on validation sets or set
by authors' recommendation. 
\mLMNN use 3
target neighbors and all imposters, while these are set to 3 and 10 in \PLML and \ourslocal. The number of anchor points in
\PLML is set to 20 as done by the authors.
For \ourslocal, we use the same basis set
as \oursglobal, and embedding dimension $D'$ is set to 40 for Vehicle, Vowel, Segment and BBC, and 100
for Letters and USPS.

\begin{figure}[t]
\centering
\subfigure[Class membership]{
   \includegraphics[width=0.25\textwidth] {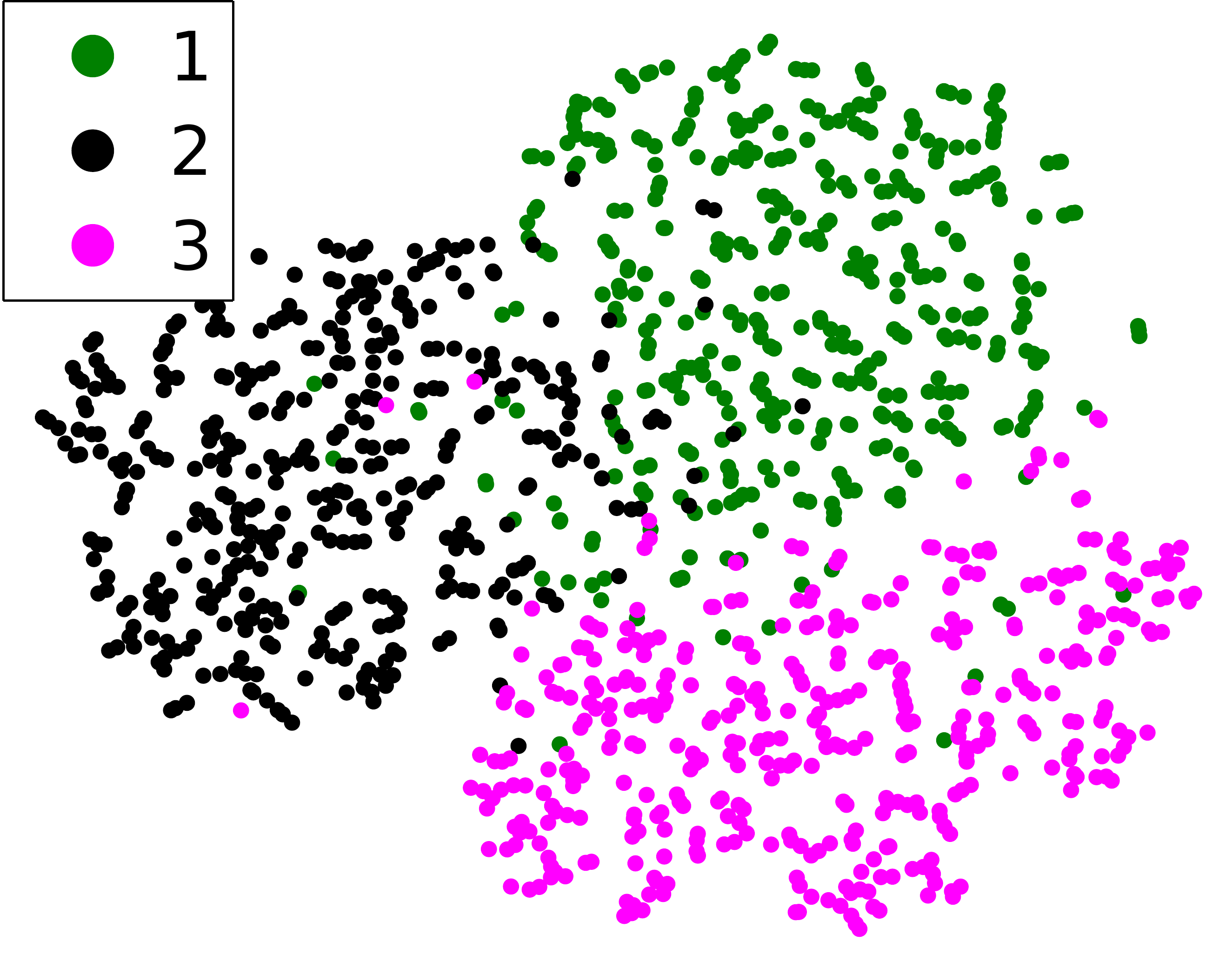}
   \label{fig:viscls}
 }\subfigure[Trained metrics]{
   \includegraphics[width=0.25\textwidth] {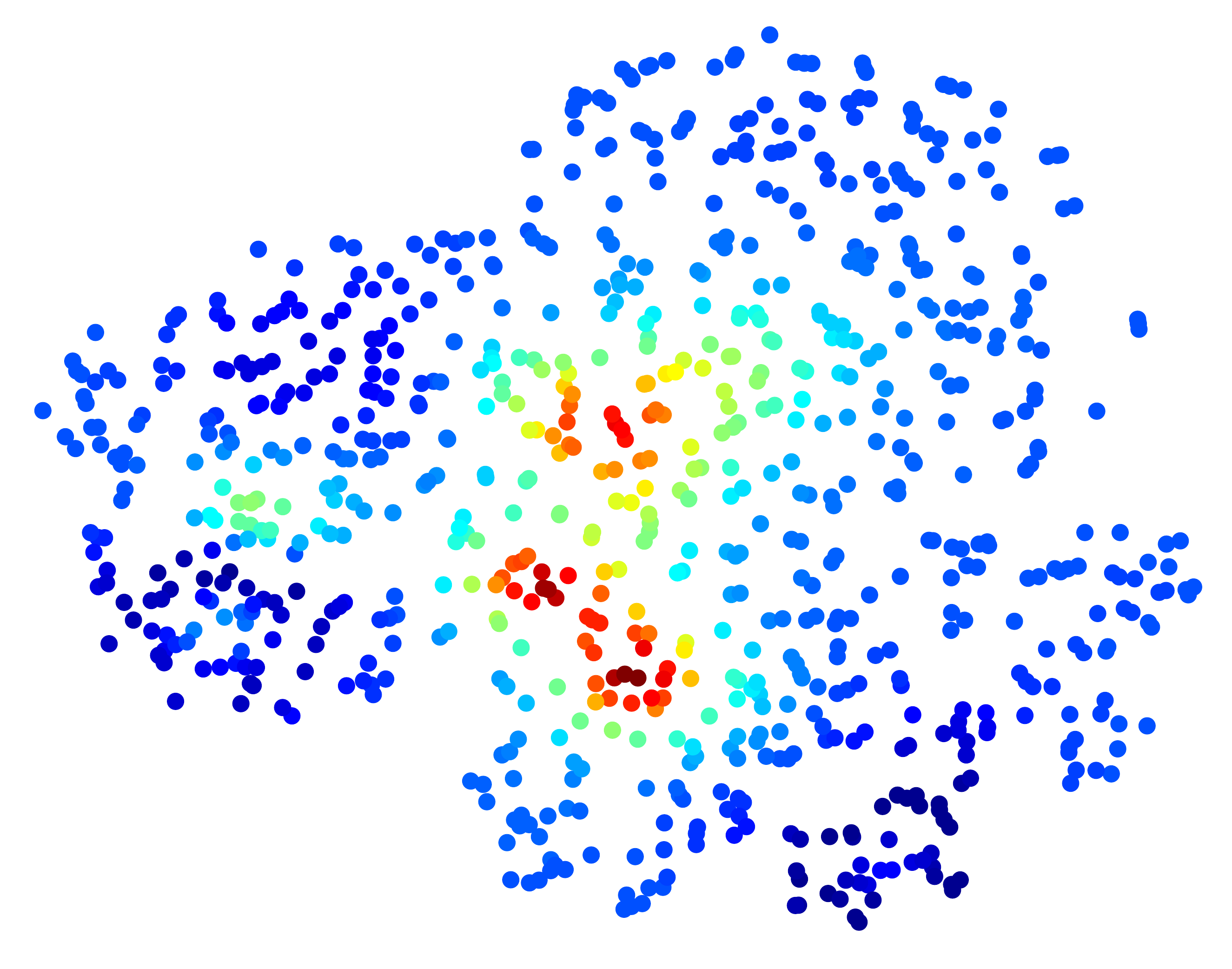}
   \label{fig:vistr}
 }\subfigure[Test metrics]{
   \includegraphics[width=0.25\textwidth] {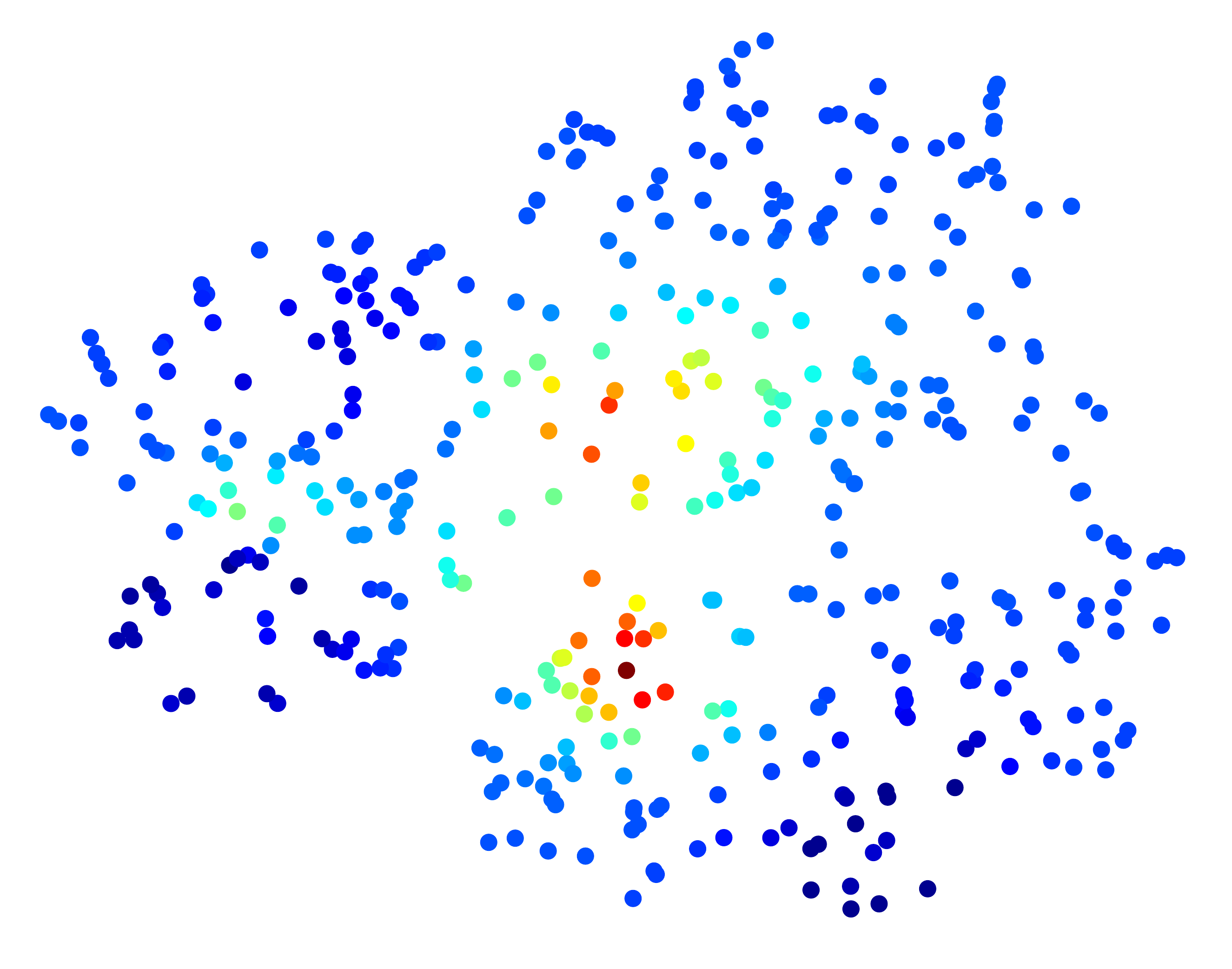}
   \label{fig:viste}
 }
\caption{Illustrative experiment on digits 1, 2 and 3 of USPS in 2D.
Refer to the main text for details.} \label{fig:vis}
\end{figure}

\begin{figure}[t]
\centering
\includegraphics[width=0.45\textwidth]{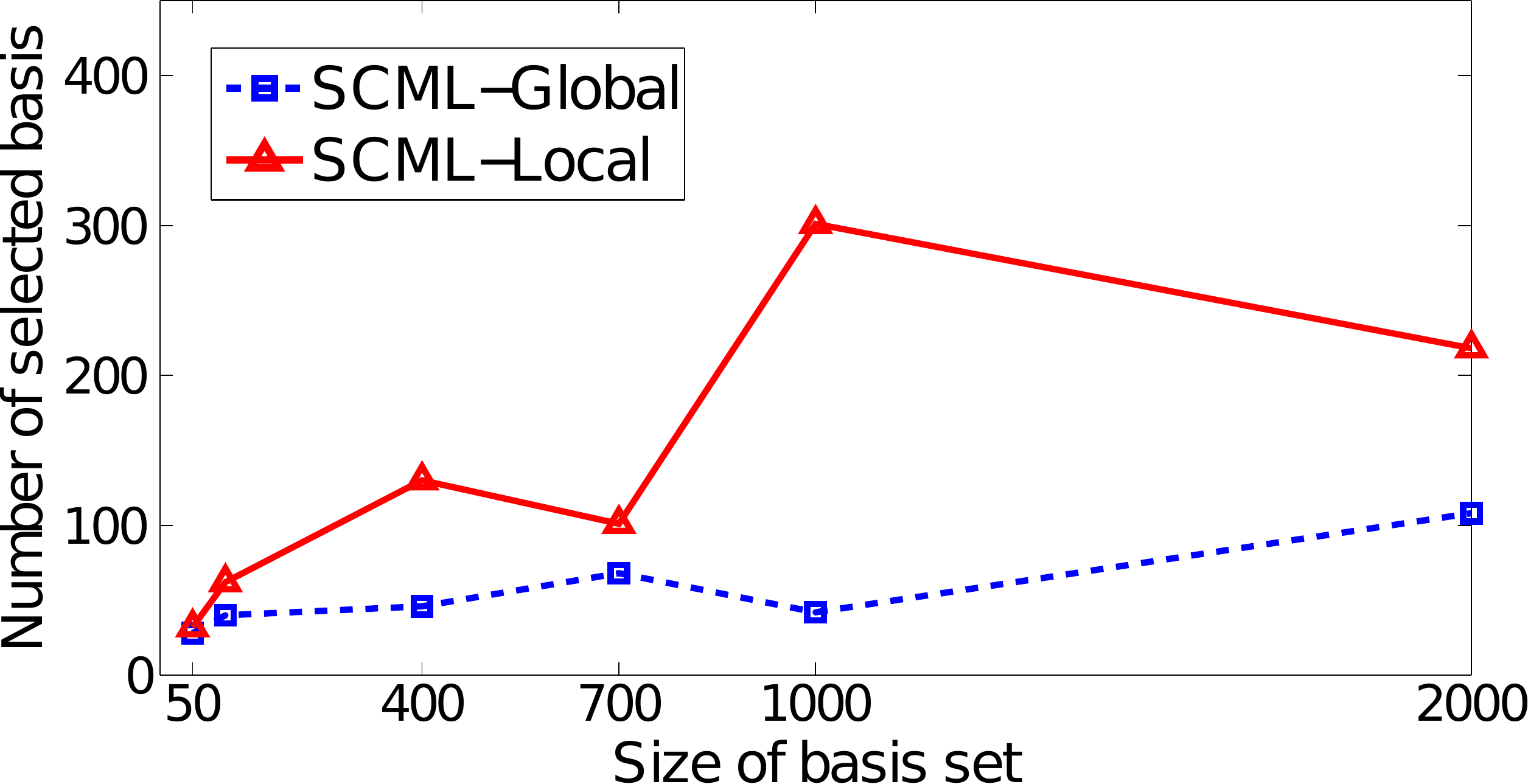}\includegraphics[width=0.45\textwidth]{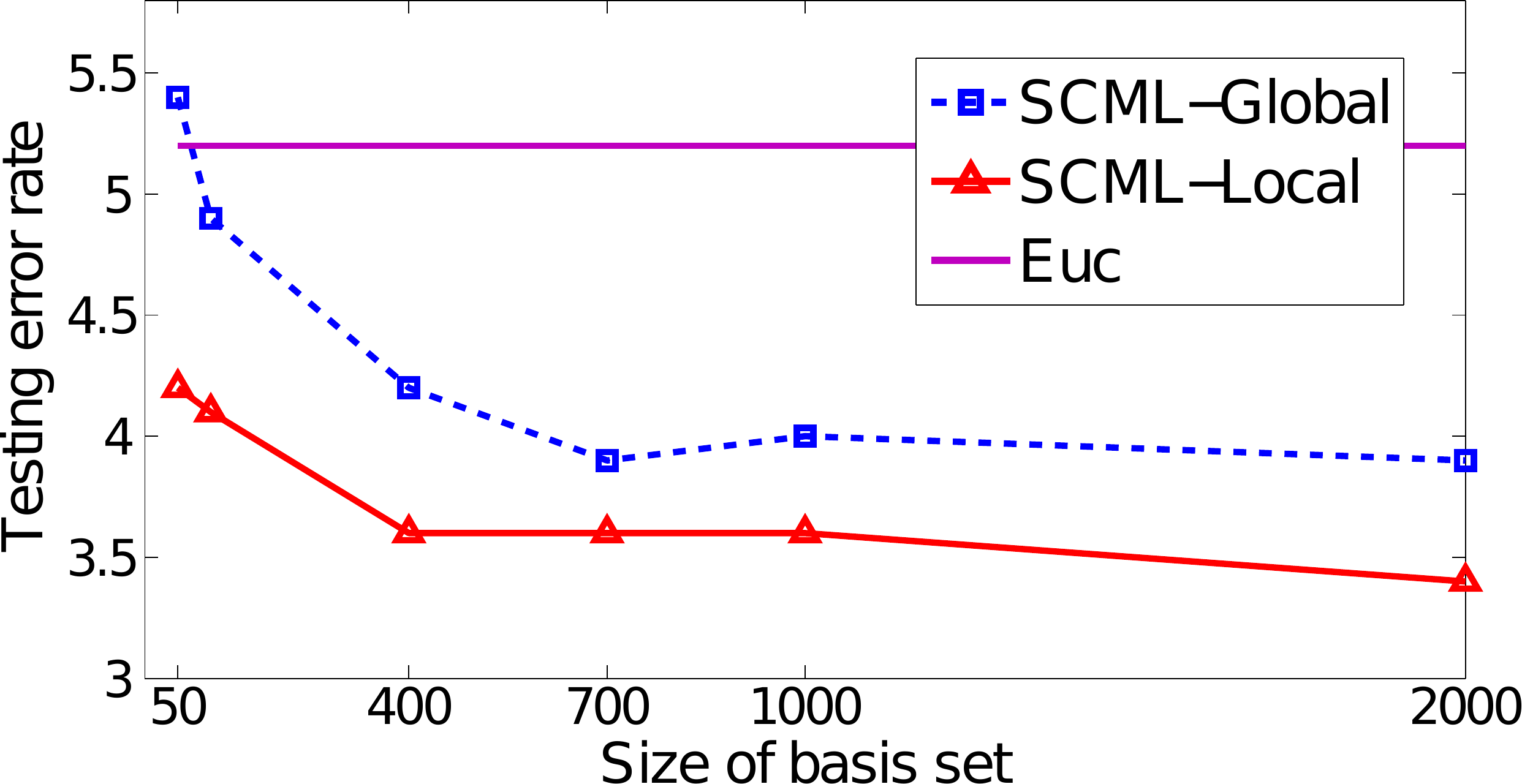}
\caption{Effect of the number of bases on Segment dataset.}
\label{fig:fNumBasis}
\end{figure}

\paragraph{Results}  Table~\ref{tab:local} gives the error rates along with the average rank of each method across all datasets. 
Note that \ourslocal significantly improves upon \oursglobal on all but one dataset and achieves the best average rank. \PLML does not perform well on small datasets (Vehicle and
Vowel), presumably because there are not enough points to get a good
estimation of the data manifold. \GLML is fast but has rather poor performance on most
datasets because its Gaussian assumption is restrictive and it
learns the local metrics independently. Among discriminative methods, \ourslocal offers the best training time, especially for high-dimensional data (e.g. on BBC, it trained in about 8 minutes, which is about 5x faster than \mLMNN and 15x faster than \PLML).
Note that on this dataset, both \mLMNN and \PLML perform
worse than \oursglobal due to severe
overfitting, while \ourslocal avoids it by learning
significantly fewer parameters. Finally, \ourslocal achieves accuracy results that are very competitive with those of a kernel SVM, as shown in Appendix~\ref{app:svm}.

\paragraph{Visualization of the learned metrics}  To provide a better understanding of why \ourslocal works well, we apply it to digits 1, 2, and 3 of USPS projected in 2D using t-SNE \citep{Maaten2008}, shown in Figure~\ref{fig:viscls}. We use 10 basis elements and $D'=5$. Figure~\ref{fig:vistr} shows the training points colored by their learned metric (based on the projection of the weight vectors in 1D using PCA). We see that the local metrics vary smoothly and are thereby robust to outliers. Unlike MM-LMNN, points within a class are allowed to have different metrics: in particular, this is useful for points that are near the decision boundary. While smooth, the variation in the weights is thus driven by discriminative information, unlike PLML where they are only based on the smoothness assumption. Finally, Figure~\ref{fig:viste} shows that the metrics consistently generalize to test data.

\paragraph{Effect of the basis set size}  Figure~\ref{fig:fNumBasis} shows the number of selected basis elements and test error rate for
\oursglobal and \ourslocal as a function of the size of basis set on Segment (results were consistent on other datasets). The left pane shows that the number of selected
elements increases sublinearly and eventually converges, while the
right pane shows that test error may be further reduced by using a
larger basis set without significant overfitting, as suggested by
our generalization bound (Theorem~\ref{thm:bound}).
Figure~\ref{fig:fNumBasis} also shows that \ourslocal generally
selects more basis elements than \oursglobal, but notice that it can
outperform \oursglobal even when the basis set is very small.

%

\section{Conclusion}
\label{sec:disc}

We proposed to learn metrics as sparse combinations of rank-one basis elements. This framework unifies several paradigms in metric learning, including global, local and multi-task learning. Of particular interest is our local metric learning algorithm which can compute instance-specific metrics for both training and test points in a principled way. The soundness of our approach is supported theoretically by a generalization bound, and we showed in experimental studies that the proposed methods improve upon state-of-the-art algorithms in terms of accuracy and scalability.



\paragraph{Acknowledgements}

This research is partially supported by  the
IARPA via DoD/ARL contract \# W911NF-12-C-0012 and DARPA via contract \# D11AP00278. The U.S. Government is
authorized to reproduce and distribute reprints for Governmental
purposes notwithstanding any copyright annotation thereon. The views and
conclusions contained herein are those of the authors and should not be
interpreted as necessarily representing the official policies or
endorsements, either expressed or implied, of IARPA, DoD/ARL, DARPA, or the
U.S. Government.

\bibliography{arXiv_metric}

\begin{thebibliography}{32}
\providecommand{\natexlab}[1]{#1}
\providecommand{\url}[1]{\texttt{#1}}
\expandafter\ifx\csname urlstyle\endcsname\relax
  \providecommand{\doi}[1]{doi: #1}\else
  \providecommand{\doi}{doi: \begingroup \urlstyle{rm}\Url}\fi

\bibitem[Argyriou et~al.(2008)Argyriou, Evgeniou, and Pontil]{Argyriou2008}
A.~Argyriou, T.~Evgeniou, and M.~Pontil.
\newblock {C}onvex multi-task feature learning.
\newblock \emph{{M}ach. {L}earn.}, 73\penalty0 (3):\penalty0 243--272, 2008.

\bibitem[Bellet and Habrard(2012)]{Bellet2012b}
A.~Bellet and A.~Habrard.
\newblock {R}obustness and {G}eneralization for {M}etric {L}earning.
\newblock Technical report, arXiv:1209.1086, 2012.

\bibitem[Bellet et~al.(2013)Bellet, Habrard, and Sebban]{Bellet2013}
Aur\'elien Bellet, Amaury Habrard, and Marc Sebban.
\newblock {A} {S}urvey on {M}etric {L}earning for {F}eature {V}ectors and
  {S}tructured {D}ata.
\newblock Technical report, arXiv:1306.6709, June 2013.

\bibitem[Blitzer et~al.(2007)Blitzer, Dredze, and Pereira]{Blitzer2007}
J.~Blitzer, M.~Dredze, and F.~Pereira.
\newblock {B}iographies, {B}ollywood, {B}oom-boxes and {B}lenders: {D}omain
  {A}daptation for {S}entiment {C}lassification.
\newblock In \emph{ACL}, 2007.

\bibitem[Caruana(1997)]{Caruana1997}
R.~Caruana.
\newblock {M}ultitask {L}earning.
\newblock \emph{{M}ach. {L}earn.}, 28\penalty0 (1):\penalty0 41--75, 1997.

\bibitem[Chang and Lin(2011)]{Chang2011}
C.-C. Chang and C.-J. Lin.
\newblock {LIBSVM} : a library for support vector machines.
\newblock \emph{{ACM} {T}ransactions on {I}ntelligent {S}ystems and
  {T}echnology}, 2\penalty0 (3):\penalty0 27--27, 2011.

\bibitem[Davis et~al.(2007)Davis, Kulis, Jain, Sra, and Dhillon]{Davis2007}
J.~Davis, B.~Kulis, P.~Jain, S.~Sra, and I.~Dhillon.
\newblock {I}nformation-theoretic metric learning.
\newblock In \emph{ICML}, 2007.

\bibitem[Duchi and Singer(2009)]{Duchi2009a}
J.~Duchi and Y.~Singer.
\newblock {E}fficient {O}nline and {B}atch {L}earning {U}sing {F}orward
  {B}ackward {S}plitting.
\newblock \emph{JMLR}, 10:\penalty0 2899--2934, 2009.

\bibitem[Frome et~al.(2007)Frome, Singer, Sha, and Malik]{Frome2007}
A.~Frome, Y.~Singer, F.~Sha, and J.~Malik.
\newblock {L}earning {G}lobally-{C}onsistent {L}ocal {D}istance {F}unctions for
  {S}hape-{B}ased {I}mage {R}etrieval and {C}lassification.
\newblock In \emph{ICCV}, 2007.

\bibitem[Goldberger et~al.(2004)Goldberger, Roweis, Hinton, and
  Salakhutdinov]{Goldberger2004}
J.~Goldberger, S.~Roweis, G.~Hinton, and R.~Salakhutdinov.
\newblock {N}eighbourhood {C}omponents {A}nalysis.
\newblock In \emph{NIPS}, 2004.

\bibitem[Gong et~al.(2012)Gong, Ye, and Zhang]{Gong2012}
P.~Gong, J.~Ye, and C.~Zhang.
\newblock {R}obust multi-task feature learning.
\newblock In \emph{KDD}, 2012.

\bibitem[Hauberg et~al.(2012)Hauberg, Freifeld, and Black]{Hauberg2012}
S.~Hauberg, O.~Freifeld, and M.~Black.
\newblock {A} {G}eometric take on {M}etric {L}earning.
\newblock In \emph{NIPS}, 2012.

\bibitem[Hong et~al.(2011)Hong, Li, Jiang, and Tu]{Hong2011}
Y.~Hong, Q.~Li, J.~Jiang, and Z.~Tu.
\newblock {L}earning a mixture of sparse distance metrics for classification
  and dimensionality reduction.
\newblock In \emph{CVPR}, 2011.

\bibitem[Jain et~al.(2008)Jain, Kulis, Dhillon, and Grauman]{Jain2008}
P.~Jain, B.~Kulis, I.~Dhillon, and K.~Grauman.
\newblock {O}nline {M}etric {L}earning and {F}ast {S}imilarity {S}earch.
\newblock In \emph{NIPS}, 2008.

\bibitem[Kolmogorov and Tikhomirov(1961)]{Kolmogorov1961}
A.~Kolmogorov and V.~Tikhomirov.
\newblock $\epsilon$-entropy and $\epsilon$-capacity of sets in functional
  spaces.
\newblock \emph{{A}merican {M}athematical {S}ociety {T}ranslations}, 2\penalty0
  (17):\penalty0 277--364, 1961.

\bibitem[Kulis(2012)]{Kulis2012}
Brian Kulis.
\newblock {M}etric {L}earning: {A} {S}urvey.
\newblock \emph{{F}oundations and {T}rends in {M}achine {L}earning}, 5\penalty0
  (4):\penalty0 287--364, 2012.

\bibitem[Noh et~al.(2010)Noh, Zhang, and Lee]{Noh2010}
Y.-K. Noh, B.-T. Zhang, and D.~Lee.
\newblock {G}enerative {L}ocal {M}etric {L}earning for {N}earest {N}eighbor
  {C}lassification.
\newblock In \emph{NIPS}, 2010.

\bibitem[Parameswaran and Weinberger(2010)]{Parameswaran2010}
S.~Parameswaran and K.~Weinberger.
\newblock {L}arge {M}argin {M}ulti-{T}ask {M}etric {L}earning.
\newblock In \emph{NIPS}, 2010.

\bibitem[Ramanan and Baker(2011)]{Ramanan2011}
D.~Ramanan and S.~Baker.
\newblock {L}ocal {D}istance {F}unctions: {A} {T}axonomy, {N}ew {A}lgorithms,
  and an {E}valuation.
\newblock \emph{TPAMI}, 33\penalty0 (4):\penalty0 794--806, 2011.

\bibitem[Sch\"olkopf et~al.(1998)Sch\"olkopf, Smola, and
  M\"uller]{Scholkopf1998}
B.~Sch\"olkopf, A.~Smola, and K.-R. M\"uller.
\newblock {N}onlinear component analysis as a kernel eigenvalue problem.
\newblock \emph{{N}eural {C}omput.}, 10\penalty0 (1):\penalty0 1299--1319,
  1998.

\bibitem[Shen et~al.(2012)Shen, Kim, Wang, and van~den Hengel]{Shen2012}
C.~Shen, J.~Kim, L.~Wang, and A.~van~den Hengel.
\newblock {P}ositive {S}emidefinite {M}etric {L}earning {U}sing {B}oosting-like
  {A}lgorithms.
\newblock \emph{JMLR}, 13:\penalty0 1007--1036, 2012.

\bibitem[Shi et~al.(2014)Shi, Bellet, and Sha]{Shi2014}
Y.~Shi, A.~Bellet, and F.~Sha.
\newblock {S}parse {C}ompositional {M}etric {L}earning.
\newblock In \emph{AAAI}, 2014.

\bibitem[van~der Maaten and Hinton(2008)]{Maaten2008}
L.~van~der Maaten and G.~Hinton.
\newblock {V}isualizing {D}ata using t-{SNE}.
\newblock \emph{JMLR}, 9:\penalty0 2579--2605, 2008.

\bibitem[Wang et~al.(2012)Wang, Woznica, and Kalousis]{Wang2012b}
J.~Wang, A.~Woznica, and A.~Kalousis.
\newblock {P}arametric {L}ocal {M}etric {L}earning for {N}earest {N}eighbor
  {C}lassification.
\newblock In \emph{NIPS}, 2012.

\bibitem[Weinberger and Saul(2009)]{Weinberger2009}
K.~Weinberger and L.~Saul.
\newblock {D}istance {M}etric {L}earning for {L}arge {M}argin {N}earest
  {N}eighbor {C}lassification.
\newblock \emph{JMLR}, 10:\penalty0 207--244, 2009.

\bibitem[Xiao(2010)]{Xiao2010}
L.~Xiao.
\newblock {D}ual {A}veraging {M}ethods for {R}egularized {S}tochastic
  {L}earning and {O}nline {O}ptimization.
\newblock \emph{JMLR}, 11:\penalty0 2543--2596, 2010.

\bibitem[Xing et~al.(2002)Xing, Ng, Jordan, and Russell]{Xing2002}
E.~Xing, A.~Ng, M.~Jordan, and S.~Russell.
\newblock {D}istance {M}etric {L}earning with {A}pplication to {C}lustering
  with {S}ide-{I}nformation.
\newblock In \emph{NIPS}, 2002.

\bibitem[Xu and Mannor(2012)]{Xu2012a}
H.~Xu and S.~Mannor.
\newblock {R}obustness and {G}eneralization.
\newblock \emph{{M}ach. {L}earn.}, 86\penalty0 (3):\penalty0 391--423, 2012.

\bibitem[Yang et~al.(2009)Yang, Kim, and Xing]{Yang2009}
X.~Yang, S.~Kim, and E.~Xing.
\newblock {H}eterogeneous multitask learning with joint sparsity constraints.
\newblock In \emph{NIPS}, 2009.

\bibitem[Ying and Li(2012)]{Ying2012}
Y.~Ying and P.~Li.
\newblock {D}istance {M}etric {L}earning with {E}igenvalue {O}ptimization.
\newblock \emph{JMLR}, 13:\penalty0 1--26, 2012.

\bibitem[Yuan and Lin(2006)]{Yuan2006}
M.~Yuan and Y.~Lin.
\newblock {M}odel selection and estimation in regression with grouped
  variables.
\newblock \emph{J. Roy. Statist. Soc. Ser. B}, 68:\penalty0 49--67, 2006.

\bibitem[Zhan et~al.(2009)Zhan, Li, Li, and Zhou]{Zhan2009}
D.-C. Zhan, M.~Li, Y.-F. Li, and Z.-H. Zhou.
\newblock {L}earning instance specific distances using metric propagation.
\newblock In \emph{ICML}, 2009.

\end{thebibliography}
\bibliographystyle{plainnat}

\begin{appendices}

\section{Detailed Analysis}
\label{app:analysis}

In this section, we give the details of the derivation of the generalization bounds for the global and multi-task learning formulations given in Section~\ref{sec:gen}.

\subsection{Preliminaries}

We start by introducing some notation.
We are given a training sample $S = \{\vct{z}_i = (\vct{x}_i,y_i)\}_{i=1}^n$ drawn i.i.d. from a distribution $P$ over the labeled space $\mathcal{Z} = \mathcal{X}\times\mathcal{Y}$. We assume that $\|\vct{x}\| \leq R$ (for some convenient norm), $\forall \vct{x}\in \mathcal{X}$. We call a triplet $(\vct{z},\vct{z}',\vct{z}'')$ \textit{admissible} if $y = y' \neq y''$. Let $C_S$ be the set of all admissible triplets built from instances in $S$.\footnote{When the training triplets consist of only a subset of all admissible triplets (which is often the case in practice), a relaxed version of the robustness property can be used to derive similar results \citep{Bellet2012b}. For simplicity, we focus here on the case when all admissible triplets are used.}

Let $L(h,\vct{z},\vct{z}',\vct{z}'')$ be the loss suffered by some hypothesis $h$ on triplet $(\vct{z},\vct{z}',\vct{z}'')$, with the convention that $L$ returns 0 for non-admissible triplets. We assume $L$ to be uniformly upper-bounded by a constant $U$. The \textit{empirical loss} $\mathcal{R}_{emp}^{C_S}(h)$ of $h$ on $C_S$ is defined as
$$\mathcal{R}_{emp}^{C_S}(h) = \frac{1}{|C_S|}\sum_{(\vct{z},\vct{z}',\vct{z}'')\in C_S} L(h,\vct{z},\vct{z}',\vct{z}''),$$
and its \textit{expected loss} $\mathcal{R}(h)$ over distribution $P$ as
$$\mathcal{R}(h) = \mathbb{E}_{\vct{z},\vct{z}',\vct{z}''\sim P} L(h,\vct{z},\vct{z}',\vct{z}'').$$
Our goal is to bound the deviation between $\mathcal{R}(\mathcal{A}_{C_S})$ and $\mathcal{R}_{emp}^{C_S}(\mathcal{A}_{C_S})$, where $\mathcal{A}_{C_S}$ is the hypothesis learned by algorithm $\mathcal{A}$ on $C_S$.

\subsection{Algorithmic Robustness}

To derive our generalization bounds, we use the recent framework of algorithmic robustness \citep{Xu2012a}, in particular its adaptation to pairwise and tripletwise loss functions used in metric learning \citep{Bellet2012b}. For the reader's convenience, we briefly review these main results below.

Algorithmic robustness is the ability of an algorithm to perform ``similarly'' on a training example and on a test example that are ``close''.
The proximity of points is based on a partitioning of the space $\mathcal{Z}$: two examples are close to each other if they lie in the same region. The partition is based on the notion of covering number \citep{Kolmogorov1961}.

\begin{definition}[Covering number]
\label{def:cover}
For a metric space $(\mathcal{S},\rho)$ and $\mathcal{V}\subset \mathcal{S}$, we say that $\hat{\mathcal{V}}\subset \mathcal{V}$ is a $\gamma$-cover of $\mathcal{V}$ if $\forall \vct{t}\in\mathcal{V}$, $\exists \hat{\vct{t}}\in\hat{\mathcal{V}}$ such that $\rho(\vct{t},\hat{\vct{t}})\leq \gamma$. The $\gamma$-covering number of $\mathcal{V}$ is
$$\mathcal{N}(\gamma,\mathcal{V},\rho) = \min\left\{|\hat{\mathcal{V}}| : \hat{\mathcal{V}}\text{ is a }\gamma\text{-cover of }\mathcal{V}\right\}.$$
\end{definition}

In particular, when $\mathcal{X}$ is compact, $\mathcal{N}(\gamma,\mathcal{X},\rho)$ is finite, leading to a finite cover.
Then, $\mathcal{Z}$ can be partitioned into $|\mathcal{Y}|\mathcal{N}(\gamma,\mathcal{X},\rho)$ subsets such that if two examples $\vct{z}=(\vct{x},y)$ and $\vct{z}'=(\vct{x}',y')$ belong to the same subset, then $y=y'$ and $\rho(\vct{x},\vct{x}')\leq\gamma$.
The definition of robustness for tripletwise loss functions \citep[adapted from][]{Xu2012a} is as follows.

\begin{definition}[Robustness for metric learning \citep{Bellet2012b}]
\label{def:robustness}
An algorithm $\mathcal{A}$ is $(N,\epsilon(\cdot))$ robust for
$N\in\mathbb{N}$ and $\epsilon(\cdot): (\mathcal{Z}\times\mathcal{Z})^n \rightarrow
\mathbb{R}$ if $\mathcal{Z}$ can be partitioned into $N$ disjoints sets, denoted
by $\{Q_i\}_{i=1}^N$, such that the following holds for all $S\in\mathcal{Z}^n$:\\
$\forall (\vct{z}_1,\vct{z}_2,\vct{z}_3) \in C_S, \forall \vct{z},\vct{z}',\vct{z}'' \in \mathcal{Z}, \forall i,j\in[N]:$  if 
$\vct{z}_1,\vct{z}\in Q_i$,  $\vct{z}_2,\vct{z}'\in Q_j$, $\vct{z}_3,\vct{z}''\in Q_k$ then 
$$|L(\mathcal{A}_{C_S},\vct{z}_1,\vct{z}_2,\vct{z}_3)-L(\mathcal{A}_{C_S},\vct{z},\vct{z}',\vct{z}'')|\leq \epsilon(C_S),$$
where $\mathcal{A}_{C_S}$ is the hypothesis learned by $\mathcal{A}$ on $C_S$.
\end{definition}

$N$ and $\epsilon(\cdot)$ quantify the robustness of the algorithm and depend on the training sample. Again adapting the result from \citep{Xu2012a}, \citep{Bellet2012b} showed that a metric learning algorithm that satisfies Definition~\ref{def:robustness} has the following generalization guarantees.

\begin{theorem}\label{thm:robustness}
If a learning algorithm $\mathcal{A}$ is $(N,\epsilon(\cdot))$-robust
and the training sample consists of the triplets $C_S$ obtained from a sample $S$ generated by $n$ i.i.d. draws from $P$, then
for any $\delta>0$, with probability at least $1-\delta$ we have:
$$
|\mathcal{R}(\mathcal{A}_{C_S})-\mathcal{R}_{emp}^{C_S}(\mathcal{A}_{C_S})|\leq \epsilon(C_S)+3U\sqrt{\frac{N \ln 2 + \ln \frac{1}{\delta}}{0.5n}}.
$$
\end{theorem}

As shown in \citep{Bellet2012b}, establishing the robustness of an algorithm is easier using the following theorem, which basically says that if a metric learning algorithm has approximately the same loss for triplets that are close to each other, then it is robust.

\begin{theorem}
Fix $\gamma>0$ and a metric $\rho$ of $\mathcal{Z}$. Suppose that $\forall \vct{z}_1,\vct{z}_2,\vct{z}_3,\vct{z},\vct{z}',\vct{z}'' : (\vct{z}_1,\vct{z}_2,\vct{z}_3)\in C_S, \rho(\vct{z}_1,\vct{z})\leq \gamma, \rho(\vct{z}_2,\vct{z}')\leq \gamma, \rho(\vct{z}_3,\vct{z}'')\leq \gamma$, $\mathcal{A}$ satisfies
$$|L(\mathcal{A}_{C_S},\vct{z}_1,\vct{z}_2,\vct{z}_3)-L(\mathcal{A}_{C_S},\vct{z},\vct{z}',\vct{z}'')|\leq \epsilon(C_S),$$
and $\mathcal{N}(\gamma/2,\mathcal{Z},\rho) < \infty$. Then the algorithm $\mathcal{A}$ is $(\mathcal{N}(\gamma/2,\mathcal{Z},\rho),\epsilon(C_S))$-robust.
\label{thm:test}
\end{theorem}

We now have all the tools we need to prove the results of interest.

\subsection{Generalization Bounds for SCML}

\subsubsection{Bound for SCML-Global}

We first focus on SCML-Global where the loss function is defined as follows:
$$L(\vct{w},\vct{z},\vct{z}',\vct{z}'') = \left[ 1 + d_{\vct{w}}(\vct{x},\vct{x}') - d_{\vct{w}}(\vct{x},\vct{x}'')\right]_+.$$
We obtain a generalization bound by showing that SCML-Global satisfies Definition~\ref{def:robustness} using Theorem~\ref{thm:test}. To establish the result, we need a bound on the $\ell_2$ norm of the basis elements. Since they are obtained by eigenvalue decomposition, their norm is equal to (and thus bounded by) 1.

Let $\vct{w}^*$ be the optimal solution to SCML-Global. By optimality of $\vct{w}^*$ we have:
$$L(\vct{w}^*,\vct{z},\vct{z}',\vct{z}'') + \beta\|\vct{w}^*\|_1 \leq L(\vct{0},\vct{z},\vct{z}',\vct{z}'') + \beta\|\vct{0}\|_1 = 1,$$
thus we get $\|\vct{w}^*\|_1 \leq 1/\beta$. Let $\mat{M}^* = \sum_{i=1}^Kw^*_i\vct{b}_i\vct{b}_i\T$ be the learned metric. Then using Holder's inequality and the bound on $\vct{w}^*$ and the $\vct{b}$'s:

$$\|\mat{M}^*\|_1 = \left\|\sum_{i=1}^Kw^*_i\vct{b}_i\vct{b}_i\T\right\|_1 = \left\|\sum_{i : w_i \neq 0}w^*_i\vct{b}_i\vct{b}_i\T\right\|_1 \leq \|\vct{w}^*\|_1\sum_{i : w_i \neq 0}\|\vct{b}_i\|_\infty\|\vct{b}_i\|_\infty \leq K^*/\beta,$$
where $K^*\leq K$ is the number of nonzero entries in $\vct{w}^*$.

Using Definition~\ref{def:cover}, we can partition $\mathcal{Z}$ into $|\mathcal{Y}|\mathcal{N}(\gamma,\mathcal{X},\rho)$ subsets such that if two examples $\vct{z}=(\vct{x},y)$ and $\vct{z}'=(\vct{x}',y')$ belong to the same subset, then $y=y'$ and $\rho(\vct{x},\vct{x}')\leq\gamma$.
Now, for $\vct{z}_1,\vct{z}_2,\vct{z}_3,\vct{z}_1',\vct{z}_2',\vct{z}_3'\in Z$, if $y_1 = y_1'$, $\|\vct{x}_1-\vct{x}_1'\|_1\leq \gamma$, $y_2 = y_2'$, $\|\vct{x}_2-\vct{x}_2'\|_1\leq \gamma$, $y_3 = y_3'$, $\|\vct{x}_3-\vct{x}_3'\|_1\leq \gamma$, then $(\vct{z}_1,\vct{z}_2,\vct{z}_3)$ and $(\vct{z}_1',\vct{z}_2',\vct{z}_3')$ are either both admissible or both non-admissible triplets.

In the non-admissible case, by definition their respective loss is 0 and so is the deviation between the losses. In the admissible case we have:
\begin{footnotesize}
\begin{eqnarray*}
&& \left|\left[1+d_{\vct{w}^*}(\vct{x}_1,\vct{x}_2) - d_{\vct{w}^*}(\vct{x}_1,\vct{x}_3)\right]_+ - \left[1+d_{\vct{w}^*}(\vct{x}_1',\vct{x}_2') - d_{\vct{w}^*}(\vct{x}_1',\vct{x}_3')\right]_+\right|\\
\leq && | (\vct{x}_1-\vct{x}_2)\T\mat{M}^*(\vct{x}_1-\vct{x}_2) - (\vct{x}_1-\vct{x}_3)\T\mat{M}^*(\vct{x}_1-\vct{x}_3) + (\vct{x}_1'-\vct{x}_3')\T\mat{M}^*(\vct{x}_1'-\vct{x}_3') - (\vct{x}_1'-\vct{x}_2')\T\mat{M}^*(\vct{x}_1'-\vct{x}_2')|\\
= && | (\vct{x}_1-\vct{x}_2)\T\mat{M}^*(\vct{x}_1-\vct{x}_2) - (\vct{x}_1-\vct{x}_2)\T\mat{M}^*(\vct{x}_1'-\vct{x}_2') + (\vct{x}_1-\vct{x}_2)\T\mat{M}^*(\vct{x}_1'-\vct{x}_2') - (\vct{x}_1'-\vct{x}_2')\T\mat{M}^*(\vct{x}_1'-\vct{x}_2')\\
&& +~(\vct{x}_1'-\vct{x}_3')\T\mat{M}^*(\vct{x}_1'-\vct{x}_3') - (\vct{x}_1'-\vct{x}_3')\T\mat{M}^*(\vct{x}_1-\vct{x}_3) + (\vct{x}_1'-\vct{x}_3')\T\mat{M}^*(\vct{x}_1-\vct{x}_3) - (\vct{x}_1-\vct{x}_3)\T\mat{M}^*(\vct{x}_1-\vct{x}_3)|\\
= && | (\vct{x}_1-\vct{x}_2)\T\mat{M}^*(\vct{x}_1-\vct{x}_2 - (\vct{x}_1'-\vct{x}_2')) + (\vct{x}_1-\vct{x}_2 - (\vct{x}_1'-\vct{x}_2'))\T\mat{M}^*(\vct{x}_1'-\vct{x}_2')\\
&& +~(\vct{x}_1'-\vct{x}_3')\T\mat{M}^*(\vct{x}_1'-\vct{x}_3' - (\vct{x}_1-\vct{x}_3)) + (\vct{x}_1'-\vct{x}_3' - (\vct{x}_1-\vct{x}_3))\T\mat{M}^*(\vct{x}_1-\vct{x}_3)|\\
\leq && | (\vct{x}_1-\vct{x}_2)\T\mat{M}^*(\vct{x}_1-\vct{x}_1')| + |(\vct{x}_1-\vct{x}_2)\T\mat{M}^*(\vct{x}_2'-\vct{x}_2)| + |(\vct{x}_1-\vct{x}_1')\T\mat{M}^*(\vct{x}_1'-\vct{x}_2')|\\
&& +~ |(\vct{x}_2' - \vct{x}_2)\T\mat{M}^*(\vct{x}_1'-\vct{x}_2')| + |(\vct{x}_1'-\vct{x}_3')\T\mat{M}^*(\vct{x}_1'-\vct{x}_1)| + |(\vct{x}_1'-\vct{x}_3')\T\mat{M}^*(\vct{x}_3 -\vct{x}_3')|\\
&& +~ |(\vct{x}_1'-\vct{x}_1)\T\mat{M}^*(\vct{x}_1-\vct{x}_3)| + |(\vct{x}_3 - \vct{x}_3')\T\mat{M}^*(\vct{x}_1-\vct{x}_3)|\\
\leq && \| \vct{x}_1-\vct{x}_2\|_\infty\|\mat{M}^*\|_1\|\vct{x}_1-\vct{x}_1'\|_1 + \|\vct{x}_1-\vct{x}_2\|_\infty\|\mat{M}^*\|_1\|\vct{x}_2'-\vct{x}_2\|_1 + \|\vct{x}_1-\vct{x}_1'\|_1\|\mat{M}^*\|_1\|\vct{x}_1'-\vct{x}_2'\|_\infty\\
&& +~ \|\vct{x}_2' - \vct{x}_2\|_1\|\mat{M}^*\|_1\|\vct{x}_1'-\vct{x}_2'\|_\infty + \|\vct{x}_1'-\vct{x}_3'\|_\infty\|\mat{M}^*\|_1\|\vct{x}_1'-\vct{x}_1\|_1 + \|\vct{x}_1'-\vct{x}_3'\|_\infty\|\mat{M}^*\|_1\|\vct{x}_3 -\vct{x}_3'\|_1\\
&& +~ \|\vct{x}_1'-\vct{x}_1\|_1\|\mat{M}^*\|_1\|\vct{x}_1-\vct{x}_3\|_\infty + \|\vct{x}_3 - \vct{x}_3'\|_1\|\mat{M}^*\|_1\|\vct{x}_1-\vct{x}_3\|_\infty \\
\leq && \frac{16\gamma RK^*}{\beta},
\end{eqnarray*}
\end{footnotesize}by using the property that the hinge loss is 1-Lipschitz, Holder's inequality and bounds on the involved quantities. Thus SCML-Global is $(|Y|\mathcal{N}(\gamma,{X},\|\cdot\|_1),\frac{16\gamma RK^*}{\beta})$-robust and the generalization bound follows.

\subsubsection{Bound for mt-SCML}

In the multi-task setting, we are given a training sample $S_t = \{\vct{z}_i = (\vct{x}^t_i,y^t_i)\}_{i=1}^{n_t}$. Let $C_{S_t}$ be the set of all admissible triplets built from instances in $S_t$.

Let $\mat{W}^*$ be the optimal solution to mt-SCML. Using the same arguments as for SCML-Global, by optimality of $\mat{W}^*$ we have $\|\mat{W}^*\|_{2,1} \leq 1/\beta$. Let $\mat{M}^*_t = \sum_{i=1}^KW^*_{ti}\vct{b}_i\vct{b}_i\T$ be the learned metric for task $t$ and $\vct{W}^*_t$ be the weight vector for task $t$, corresponding to the $t$-th row of $\vct{W}^*$. Then using the fact that $\|\vct{W}^*_t\|_{2,1} \leq \|\mat{W}^*\|_{2,1}$ and $\|\vct{b}\|_{2,1} = \|\vct{b}\|_2$, we have:
$$\|\mat{M}^*_t\|_{2,1}= \left\|\sum_{i=1}^KW^*_{ti}\vct{b}_i\vct{b}_i\T\right\|_{2,1}= \left\|\sum_{i : W^*_{ti} \neq 0}W^*_{ti}\vct{b}_i\vct{b}_i\T\right\|_{2,1}\leq \|\vct{W}^*_t\|_{2,1}\sum_{i : W^*_{ti} \neq 0}\|\vct{b}_i\|_{2,1}\|\vct{b}_i\|_{2,1}\leq K_t^*/\beta,$$
where $K_t^* \leq K$ is the number of nonzero entries in $\vct{W}^*_t$.

From this we can derive a generalization bound for each task using arguments similar to the global case, using a partition specific to each task defined with respect to $\|\cdot\|_2$. Without loss of generality, we focus on task $t$ and only explicitly write the last derivations as the beginning is the same as above:
\begin{small}
\begin{eqnarray*}
&& \left|\left[1+d_{\vct{w}^*}(\vct{x}_1,\vct{x}_2) - d_{\vct{w}^*}(\vct{x}_1,\vct{x}_3)\right]_+ - \left[1+d_{\vct{w}^*}(\vct{x}_1',\vct{x}_2') - d_{\vct{w}^*}(\vct{x}_1',\vct{x}_3')\right]_+\right|\\
\leq && | (\vct{x}_1-\vct{x}_2)\T\mat{M}^*(\vct{x}_1-\vct{x}_1')| + |(\vct{x}_1-\vct{x}_2)\T\mat{M}^*(\vct{x}_2'-\vct{x}_2)| + |(\vct{x}_1-\vct{x}_1')\T\mat{M}^*(\vct{x}_1'-\vct{x}_2')|\\
&& +~ |(\vct{x}_2' - \vct{x}_2)\T\mat{M}^*(\vct{x}_1'-\vct{x}_2')| + |(\vct{x}_1'-\vct{x}_3')\T\mat{M}^*(\vct{x}_1'-\vct{x}_1)| + |(\vct{x}_1'-\vct{x}_3')\T\mat{M}^*(\vct{x}_3 -\vct{x}_3')|\\
&& +~ |(\vct{x}_1'-\vct{x}_1)\T\mat{M}^*(\vct{x}_1-\vct{x}_3)| + |(\vct{x}_3 - \vct{x}_3')\T\mat{M}^*(\vct{x}_1-\vct{x}_3)|\\
\leq && \| \vct{x}_1-\vct{x}_2\|_2\|\mat{M}^*\|_\mathcal{F}\|\vct{x}_1-\vct{x}_1'\|_2 + \|\vct{x}_1-\vct{x}_2\|_2\|\mat{M}^*\|_\mathcal{F}\|\vct{x}_2'-\vct{x}_2\|_2 + \|\vct{x}_1-\vct{x}_1'\|_2\|\mat{M}^*\|_\mathcal{F}\|\vct{x}_1'-\vct{x}_2'\|_2\\
&& +~ \|\vct{x}_2' - \vct{x}_2\|_2\|\mat{M}^*\|_\mathcal{F}\|\vct{x}_1'-\vct{x}_2'\|_2 + \|\vct{x}_1'-\vct{x}_3'\|_2\|\mat{M}^*\|_\mathcal{F}\|\vct{x}_1'-\vct{x}_1\|_2 + \|\vct{x}_1'-\vct{x}_3'\|_2\|\mat{M}^*\|_\mathcal{F}\|\vct{x}_3 -\vct{x}_3'\|_2\\
&& +~ \|\vct{x}_1'-\vct{x}_2\|_2\|\mat{M}^*\|_\mathcal{F}\|\vct{x}_1-\vct{x}_3\|_2 + \|\vct{x}_3 - \vct{x}_3'\|_2\|\mat{M}^*\|_\mathcal{F}\|\vct{x}_1-\vct{x}_3\|_2 \\
\leq && \frac{16\gamma RK_t^*}{\beta},
\end{eqnarray*}
\end{small}where we used the same arguments as above and the inequality $\|\mat{M}^*\|_{F} \leq \|\mat{M}^*\|_{2,1}$. Thus mt-SCML is $(|Y|\mathcal{N}(\gamma,{X},\|\cdot\|_2),\frac{16\gamma RK_t^*}{\beta})$-robust and the bound for task $t$ follows. Note that the number of training examples in the bound is only that from task $t$, i.e., $n = n_t$.

\subsubsection{Comments on SCML-Local}

It would be interesting to be able to derive a similar bound for SCML-Local. Unfortunately, as it is nonconvex, we cannot assume optimality of the solution. If a similar formulation can be made convex, the same proof technique should apply: even though each instance has its own metric, it essentially depends on the instance itself (whose norm is bounded) and on the learned parameters shared across metrics (which could be bounded using optimality of the solution). Deriving such a convex formulation and the corresponding generalization bound is left as future work.

\section{Experimental Comparison with Support Vector Machines}
\label{app:svm}

\begin{table}[t]
\centering
\small{ \tabcolsep=0.13cm
\begin{tabular}{|c|c|c|c|c|}
\hline Dataset & Linear SVM & Kernel SVM & \oursglobal & \ourslocal\\
\hline\hline Vehicle &21.4$\pm$0.6& \textbf{16.6$\pm$0.8} & 21.3$\pm$0.6 & \textbf{18.0$\pm$0.6}\\
\hline Vowel & 24.3$\pm$0.7 & \textbf{4.4$\pm$0.4} & 10.9$\pm$0.5 & 6.1$\pm$0.4\\
\hline Segment &5.1$\pm$0.2& \textbf{3.6$\pm$0.2} & 4.1$\pm$0.2 & \textbf{3.6$\pm$0.2}\\
\hline Letters &19.5$\pm$0.4& 8.8$\pm$0.2 & 9.0$\pm$0.2 & \textbf{8.3$\pm$0.2} \\
\hline USPS &6.5$\pm$0.2& 4.2$\pm$0.1 & 4.1$\pm$0.1 & \textbf{3.6$\pm$0.1} \\
\hline BBC & 4.3$\pm$0.2 & \textbf{3.8$\pm$0.2} & \textbf{3.9$\pm$0.2} & \textbf{4.1$\pm$0.2} \\
\hline\hline
Avg. rank & 3.8 & 1.3 & 2.3 & \textbf{1.2}\\
\hline
\end{tabular}}
\caption{Comparison of SCML against linear and kernel SVM (best in bold).}
\label{tab:svm}
\end{table}

In this section, we compare \oursglobal and \ourslocal to Support Vector Machines (SVM) using a linear and a RBF kernel. We used the software \texttt{LIBSVM} \citep{Chang2011} and tuned the parameter $C$ as well as the bandwidth for the RBF kernel on the validation set.
Table~\ref{tab:svm} shows misclassification rates averaged over 20 random splits, along with standard error and the average rank of each method across all datasets. First, we can see that \oursglobal consistently performs better than linear SVM. Second, \ourslocal is competitive with kernel SVM. These results show that a simple $k$-nearest neighbor strategy with a good metric can be competitive (and even outperform) SVM classifiers.

\end{appendices}

\end{document}